\documentclass[letterpaper, 10 pt, conference]{ieeeconf}  
\usepackage{booktabs}

\usepackage{multirow}
\usepackage{graphicx} 
\usepackage{float} 
\usepackage{amsmath} 
\usepackage{amssymb}
\usepackage{bm}

\usepackage{mathtools}
\usepackage{booktabs} 
\usepackage{tabularx} 
\usepackage{arydshln} 
\newcommand{\gaussians}{\mathcal{G}}
\usepackage{amssymb}
\usepackage{placeins} 
\usepackage{hyperref} 

\IEEEoverridecommandlockouts                              

\title{\LARGE \bf
Embracing Dynamics: Dynamics-aware 4D Gaussian Splatting SLAM
}

\author{Zhicong Sun$^{1,2}$, Jacqueline Lo* $^{1}$ and Jinxing Hu*$^{2}$
\thanks{$^{*}$ Corresponding authors: Jinxing Hu and Jacqueline Lo.}
\thanks{$^{1}$ Hong Kong Polytechnic University, Hong Kong SAR, China.}%
\thanks{$^{2}$ Shenzhen Institutes of Advanced Technology, Chinese Academy of Sciences, Shenzhen, China.}%
\thanks{$^{+}$ Emails: zhicong.sun@connect.polyu.hk, jinxing.hu@siat.ac.cn, jacqueline.lo@polyu.edu.hk}}

\begin{document}

\maketitle
\thispagestyle{empty}
\pagestyle{empty}

\begin{abstract}
Simultaneous localization and mapping (SLAM) technology has recently achieved photorealistic mapping capabilities thanks to the real-time, high-fidelity rendering enabled by 3D Gaussian Splatting (3DGS).
However, due to the static representation of scenes, current 3DGS-based SLAM encounters issues with pose drift and failure to reconstruct accurate maps in dynamic environments. To address this problem, we present D4DGS-SLAM, the first SLAM method based on 4DGS map representation for dynamic environments. By incorporating the temporal dimension into scene representation, D4DGS-SLAM enables high-quality reconstruction of dynamic scenes. Utilizing the dynamics-aware InfoModule, we can obtain the dynamics, visibility, and reliability of scene points, and filter out unstable dynamic points for tracking accordingly. When optimizing Gaussian points, we apply different isotropic regularization terms to Gaussians with varying dynamic characteristics. Experimental results on real-world dynamic scene datasets demonstrate that our method outperforms state-of-the-art approaches in both camera pose tracking and map quality.
\end{abstract}

\section{INTRODUCTION}
\begin{figure}[htbp]
\centerline{\includegraphics[width=\columnwidth]{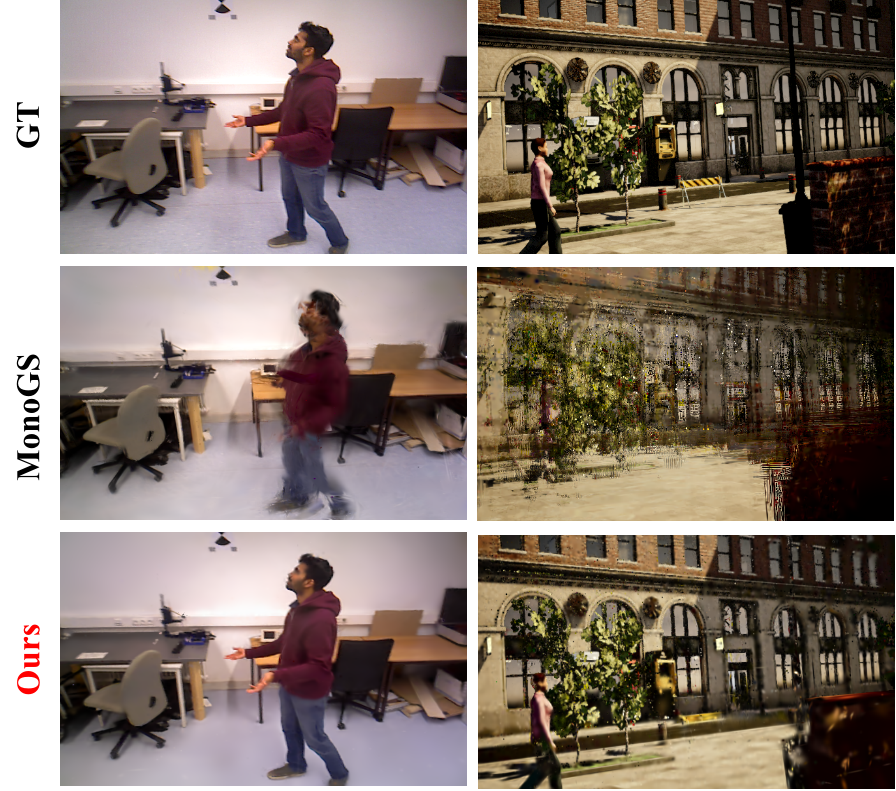}}
\caption{Comparison of rendering results between 3DGS-based MonoGS and our 4DGS-based D4DGS-SLAM on Bonn and TartanAir-Shibuya Datasets.}
\label{fig:first_view}
\end{figure}

Simultaneous Localization and Mapping (SLAM) has been a cornerstone in robotics and computer vision, enabling applications such as autonomous navigation, augmented reality, and 3D reconstruction.  
Recently, SLAM has benefited from 3D Gaussian Splatting (3DGS)~\cite{kerbl20233d}, which offers explicit map representation and real-time high-fidelity rendering.
3DGS-based methods such as SplaTAM~\cite{keetha2023splatam}, MonoGS~\cite{matsuki2023gaussian}, and GS-SLAM~\cite{yan2023gs} have achieved state-of-the-art (SOTA) map reconstruction quality and speed in static environments. 
However, these methods face challenges in dynamic environments with moving objects, leading to camera pose estimation drift and the appearance of artifacts in the map, or even to map reconstruction failure, as shown in Fig. \ref{fig:first_view}.
Recent work such as DG-SLAM~\cite{yan2023gs} improves tracking performance in dynamic environments by filtering dynamic objects. However, since it depends on a static 3DGS representation, it cannot fully reconstruct the entire scene nor produce high-quality complete maps.

Beyond 3DGS-based SLAM, SLAM methods based on Neural Radiance Fields (NeRF)~\cite{mildenhall2021nerf} also provide photorealistic maps. Following the demonstration of high-quality map reconstruction capabilities by NeRF-based SLAM methods such as iMAP~\cite{imap}, NICE-SLAM~\cite{nice_slam}, ESLAM \cite{Johari_2023_CVPR} and Co-SLAM\cite{Wang_2023_CVPR}, approaches like DN-SLAM~\cite{ruan2023dn}, NID-SLAM~\cite{xu2024nid}, DDN-SLAM~\cite{ddnslam}, and RodYN-SLAM~\cite{jiang2024rodyn} have extended NeRF's capabilities to dynamic scenes. Although these methods achieve map reconstruction quality comparable to 3DGS-based methods, their real-time performance is hindered by the significant computational demands of ray tracing used in map rendering.

To achieve accurate camera pose estimation and real-time high-fidelity map reconstruction in dynamic scenes, we propose DD4DGS-SLAM. 
To the best of our knowledge, this work presents the first SLAM system based on a dynamic map representation.
Unlike existing 3DGS-based SLAM methods that filter out dynamic objects, we choose to ``embrace" dynamics by using 4D Gaussian Splatting (4DGS)~\cite{yang2023gs4d} with a temporal dimension for map representation and rendering. 4DGS enables our system to effectively learn the spatio-temporal characteristics of the scene, thereby reconstructing a map with few artifacts without filtering out dynamic objects. 
To reduce tracking drifts in dynamic environments when using photometric and geometric error-based optimization, we introduce a long-term effective any point tracking (LEAP)~\cite{chen2024leap} for dynamics-aware tracking point filtering.
By leveraging the dynamics and reliability of anchors in the scene provided by LEAP, we can select stable and static points in the current frame for tracking, significantly improving the accuracy of camera pose estimation. 
Dynamic awareness also facilitates the densification and pruning of Gaussians over time, allowing the Gaussian set to better fit dynamic scenes.
The experimental results on real-world dynamic datasets demonstrate that our method achieves SOTA performance in both camera pose estimation and mapping quality.
Our contributions are summarized as follows:
\begin{enumerate}
    \item We propose the first SLAM system based on a 4DGS with dynamic scene representation capabilities, enabling spatio-temporal modeling of dynamic environments.
    \item We enhance the SLAM system with the dynamics-aware InfoModule, upon which we build strategies and a pipeline for tightly integrating dynamic information, significantly improving tracking and mapping capabilities.
    \item Our method achieves SOTA tracking and mapping performance on multiple dynamic SLAM benchmarks and demonstrates the capability to reconstruct complete maps of dynamic scenes under precise pose estimation.
\end{enumerate}

\section{Related Work}
Early SLAM systems laid the foundation for real-time localization and mapping through geometric feature tracking. Seminal works like ORB-SLAM2 \cite{orbslam2} and ORB-SLAM3 \cite{orbslam3} established feature-based pipelines using sparse hand-crafted descriptors. These methods excel in static environments but suffer severe degradation when dynamic objects dominate observations, as moving features introduce false geometric constraints. Subsequent improvements like DS-SLAM~\cite{ds_slam}, DynaSLAM \cite{dynaslam}, VDO-SLAM~\cite{vdoslam}, FlowFusion ~\cite{flowfusion}, Raft~\cite{teed2020raft}, Gmflow~\cite{xu2022gmflow} attempted to address this through semantic segmentation masks or optical flow, yet remained limited either by their dependency on pre-trained recognition models or inability to handle unknown moving entities. The fundamental constraint of these systems lies in their static world assumption.

The emergence of Neural Radiance Fields (NeRF)~\cite{mildenhall2021nerf} catalyzed a paradigm shift towards continuous scene representations. Pioneering neural SLAM systems like iMAP~\cite{imap}, NICE-SLAM~\cite{nice_slam}, ESLAM \cite{Johari_2023_CVPR} and Co-SLAM\cite{Wang_2023_CVPR} combined implicit geometry encoding with simultaneous pose optimization, achieving unprecedented reconstruction quality. However, their volumetric ray-marching pipelines incur heavy computational costs, making them impractical for real-time applications. More critically, these methods inherently entangle static and dynamic elements within their MLP weights, causing catastrophic mapping failures when objects move during camera observation. While recent works like DN-SLAM~\cite{ruan2023dn}, NID-SLAM~\cite{xu2024nid}, DDN-SLAM~\cite{ddnslam}, and RodYN-SLAM~\cite{jiang2024rodyn} attempt dynamic and static decomposition through optical flow or semantic segmentation, their neural rendering backbone remains too slow for interactive SLAM. This performance-realism tradeoff motivated the community to seek alternative representations.

3DGS~\cite{kerbl20233d} revolutionized scene representation through explicit, differentiable rasterization. By modeling scenes as anisotropic Gaussians with view-dependent appearance, 3DGS achieves photorealistic rendering at 200+ FPS while maintaining explicit geometric control. This breakthrough inspired Gaussian-based SLAM systems like SplaTAM~\cite{keetha2023splatam}, MonoGS \cite{matsuki2023gaussian} and GS-SLAM~\cite{yan2023gs}, which optimize Gaussian parameters alongside camera poses. Though these methods set new SOTA in static scene reconstruction, they inherit the static assumption bottleneck, namly dynamic objects cause irreversible corruption of the Gaussian map due to unmodeled temporal variations. Despite efforts by recent works such as DG-SLAM~\cite{yan2023gs} to tackle tracking issues in dynamic environments by filtering out dynamic objects, the reliance on a static 3DGS representation limits its ability to fully reconstruct an accurate map.
Unlike the aforementioned works, our approach embraces dynamic objects and learns their characteristics rather than removing them. As demonstrated below, this concept significantly enhances the capabilities of the SLAM system in dynamic environments.

\section{METHOD}
\begin{figure*}[htbp]
\vspace{1.2mm}
\centerline{\includegraphics[width=1.9\columnwidth]{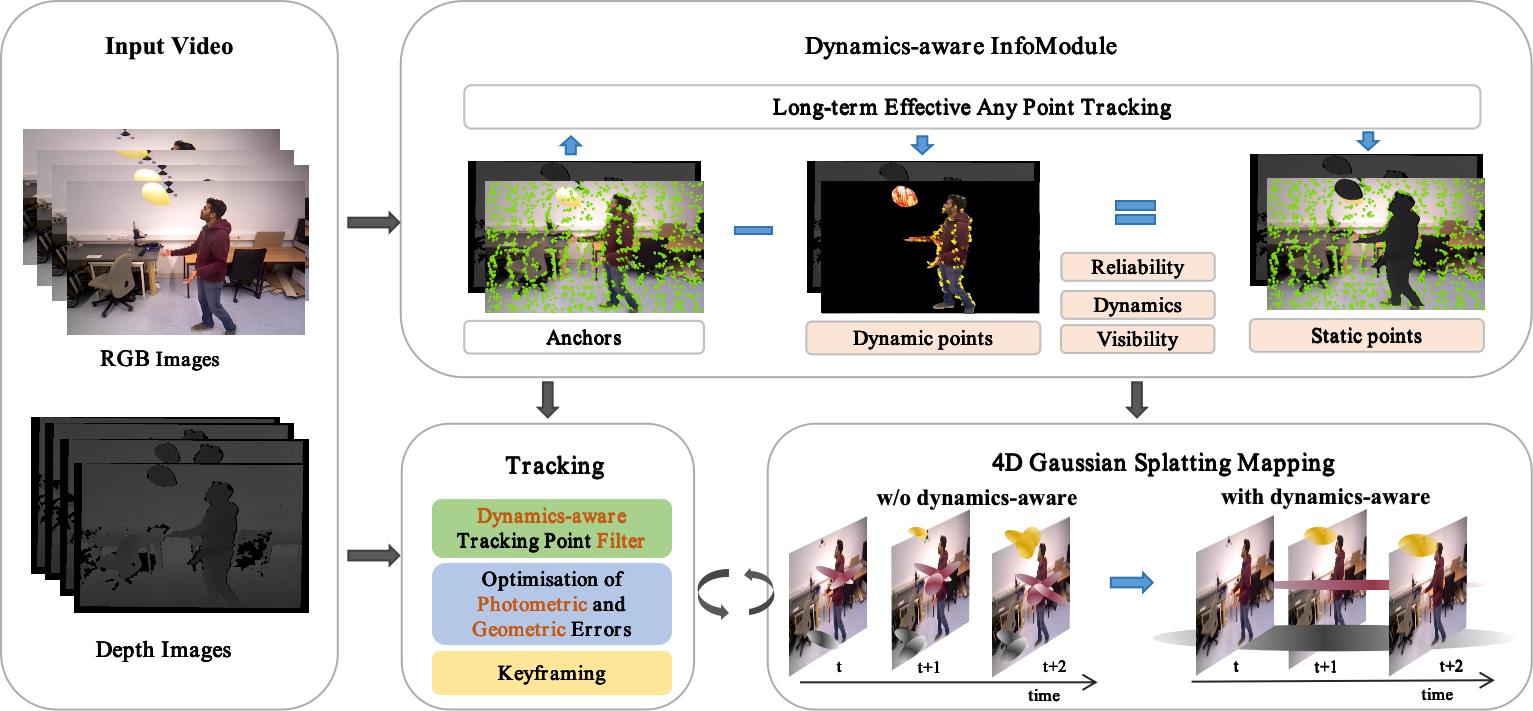}}
\caption{Overview of D4DGS-SLAM}
\label{fig:over}
\end{figure*}

The pipeline of D4DGS-SLAM is illustrated in Fig. \ref{fig:over}. Our system uses an RGB-D image sequence as input. We first extract anchors from each incoming RGB frame that are well-distributed globally and reflect the image features. These anchors, along with the RGB images, are fed into the LEAP module to obtain the dynamics and reliability of the anchors. This allows us to distinguish between stable dynamic points and static points. The static anchors are used for tracking to estimate the camera pose. These poses and dynamic information are then sent to the mapping module. We use 4DGS for mapping and select different scale penalty factors based on the dynamics and reliability of the covered points to control the distribution of the Gaussian in space-time. The techniques used and our SLAM system will be introduced below.

\subsection{4D Gaussian Map Representation}
\label{sec:4d_gaussian_map}
Our SLAM system represents a dynamic scene by a set of anisotropic 4D Gaussians. Each Gaussian captures the spatio-temporal characteristics of a region in the scene. The unnormalised probability density function of a 4D Gaussian is given by:
\begin{equation}
    p(\bm{x}|\bm{\mu}, \bm{\Sigma}) = e^{-\frac{1}{2} \left( \bm{x}-\bm{\mu} \right)^T \bm{\Sigma}^{-1} \left( \bm{x}-\bm{\mu} \right)},
\end{equation}
where $\bm{x} = (x, y, z, t)$ denotes a point in the 4D spatio-temporal space, $\bm{\mu} = (\mu_x, \mu_y, \mu_z, \mu_t)$ is the mean vector representing the Gaussian's position in space and time, and $\bm{\Sigma} \in \mathbb{R}^{4 \times 4}$ is the covariance matrix describing its shape and orientation and encoding the Gaussian's spatial extent and temporal duration. $\bm{\Sigma}$ is parameterized as $ \bm{\Sigma} = \bm{R} \bm{S} \bm{S}^T \bm{R}^T$,
where $\bm{S} = \mathrm{diag}(s_x, s_y, s_z, s_t)$ is the scaling matrix, and $\bm{R} = L(q_l) R(q_r)$ is the 4D rotation matrix constructed from two quaternions $q_l$ and $q_r$.
The 4D Gaussian can be decomposed into a conditional 3D Gaussian $p(x,y,z|t) = \mathcal{N} \left( (x,y,z); \bm{\mu}_{xyz|t}, \bm{\Sigma}_{xyz|t} \right)$ for spatial distribution and a marginal 1D Gaussian $p(t) = \mathcal{N} \left( t; \mu_{4}, \Sigma_{4,4} \right)$ for temporal distribution. The conditional mean $\bm{\mu}_{xyz|t}$ and covariance $\bm{\Sigma}_{xyz|t}$ are derived as:
\begin{equation}
\begin{aligned}
    \bm{\mu}_{xyz|t} &= \bm{\mu}_{1:3} + \bm{\Sigma}_{1:3,4}\bm{\Sigma}_{4,4}^{-1} (t - \mu_t), \\
    \bm{\Sigma}_{xyz|t} &= \bm{\Sigma}_{1:3,1:3} - \bm{\Sigma}_{1,3,4} \bm{\Sigma}_{4,4}^{-1} \bm{\Sigma}_{4,1:3}.
\end{aligned}
\end{equation}

In the map rendering process, the color of a pixel $(u, v)$ at time $t$ is computed by blending the contributions of visible 4D Gaussians. 
This rendering process ensures smooth transitions in both space and time. 
The rendering equation is formulated as:
\begin{equation}
\begin{split}
    \mathcal{I}(u,v,t) = \sum^{N}_{i=1} p_{i}(t) p_{i}(u,v|t) \alpha_i c_{i}(d,t) \\
    \times \prod^{i-1}_{j=1} (1- p_{j}(t) p_{j}(u,v|t) \alpha_j),
\end{split}
\end{equation}
Here, $p_i(t)$ is the marginal temporal distribution of the $i$-th Gaussian, derived from the 4D Gaussian's temporal component as $p_i(t) = \mathcal{N} \left( t; \mu_{t,i}, \Sigma_{t,i} \right)$. The term $p_i(u,v|t)$ is the conditional spatial distribution of the $i$-th Gaussian, obtained by projecting the 3D Gaussian $p_i(x,y,z|t)$ onto the image plane, where $p_i(u,v|t) = \mathcal{N} \left( (u,v); \bm{\mu}_{uv|t,i}, \bm{\Sigma}_{uv|t,i} \right)$. The opacity $\alpha_i$ is directly obtained from the Gaussian's optical properties, while the view-dependent color $c_i(d,t)$ is represented using 4D Spherindrical Harmonics (4DSH) as that in \cite{yang2023gs4d}. This decomposition enables the 4D Gaussian to separately model the spatial and temporal characteristics of dynamic scenes, which is crucial for the rendering process.

\subsection{Long-term Effective Any Point Tracking}
\label{sec:leap}
Our SLAM system integrates the pretrained LEAP module \cite{chen2024leap} to robustly track query points across image sequences. LEAP provides three highly beneficial pieces of information for dynamic SLAM: visibility, dynamic status, and reliability of visibility for each tracked point in the current frame.

Considering a sequence of $S$ consecutive RGB images $\mathbf{I}=[\mathbf{I}_1,...,\mathbf{I}_S]$, LEAP predicts the trajectory $\mathbf{X} = [\mathbf{x}_1, ..., \mathbf{x}_{S}]$, dynamic track label $\mathbf{M_d} = [\mathbf{m_d}_1, ..., \mathbf{m_d}_{S}]$, visibility $\mathbf{V} = [v_1,..., v_{S}]$ and reliability of visibility $\mathbf{\Phi} = [\phi_1,..., \phi_{S}]$ for a query point $\mathbf{x}_q$ in frame $s_q$. This can is expressed as:
\begin{equation}
    (\mathbf{X}, \mathbf{M_d}, \mathbf{V}, \mathbf{\Phi}) = \text{LEAP}(\mathbf{I}, \mathbf{x}_q, s_q).
\end{equation}

Visibility indicates whether a tracking point is visible or occluded in a given frame. LEAP predicts the visibility of each point throughout the sequence as a binary label \( v_s \in \{0,1\} \), where \( v_s = 1 \) indicates the point is visible in frame \( s \), and \( v_s = 0 \) indicates occlusion. The visibility is derived from the final point feature \( \mathbf{F}^K \) after \( K \) iterations of refinement, using a simple linear projection layer \( \mathcal{G}_v \):

\begin{equation}
    \mathbf{V} = \mathcal{G}_v(\mathbf{F}^K),
\end{equation}

\begin{figure}[htbp]
\centerline{\includegraphics[width=0.9\columnwidth]{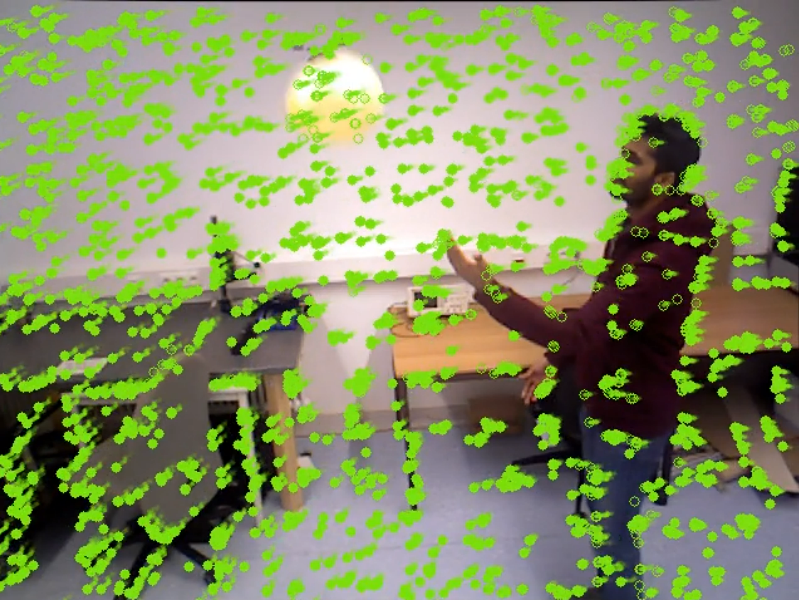}}
\caption{Visualization of anchors.}
\vspace{-2mm}
\label{fig:leap_point_anchor}
\end{figure}

Dynamic status determines whether a point belongs to a static or dynamic object in the scene. LEAP predicts the dynamic label \( \mathbf{m}_d \) for each trajectory by leveraging both visual appearance and long-term motion cues. Specifically, LEAP tracks additional anchors alongside the original queries to capture global motion patterns. The anchors are selected by the gradient-based sampling method. 
Fig. \ref{fig:leap_point_anchor} visualizes the anchor points, which are widely distributed in regions with high gradient magnitudes. These anchor points are utilized during tracking initialization and keypoint addition, ensuring robust and distinctive point selection for subsequent tracking. 
The dynamic label is estimated using a shallow MLP layer \( \mathcal{G}_d \) with average pooling over the temporal dimension:

\begin{equation}
    \mathbf{m}_d = \text{avgpool}(\mathcal{G}_d([\mathbf{X}^K; \mathbf{X}^K_A], [\mathbf{F}^K; \mathbf{F}^K_A])),
\end{equation}
where \( \mathbf{X}^K \) and \( \mathbf{F}^K \) are the final point trajectory and features, respectively, and \( \mathbf{X}^K_A \) and \( \mathbf{F}^K_A \) are the corresponding anchor trajectory and features.

Reliability of visibility quantifies the confidence in the visibility prediction. LEAP incorporates a probabilistic formulation to model the uncertainty of point trajectories. The uncertainty \( \phi(\mathbf{x}_s) \) for each point is derived from the scale matrices \( \mathbf{\Sigma}_a \) and \( \mathbf{\Sigma}_b \) of the multivariate Cauchy distribution:

\begin{equation}
    \phi(\mathbf{x}_s) = \mathbf{\Sigma}_a[s,s] + \mathbf{\Sigma}_b[s,s],
\end{equation}
where \( \mathbf{\Sigma}_a \) and \( \mathbf{\Sigma}_b \) are the scale matrices for the X and Y coordinates, respectively. Points with low uncertainty (high reliability) are prioritized for tracking and optimization.

\vspace{-2mm}
\begin{figure}[htbp]
\centerline{\includegraphics[width=0.9\columnwidth]{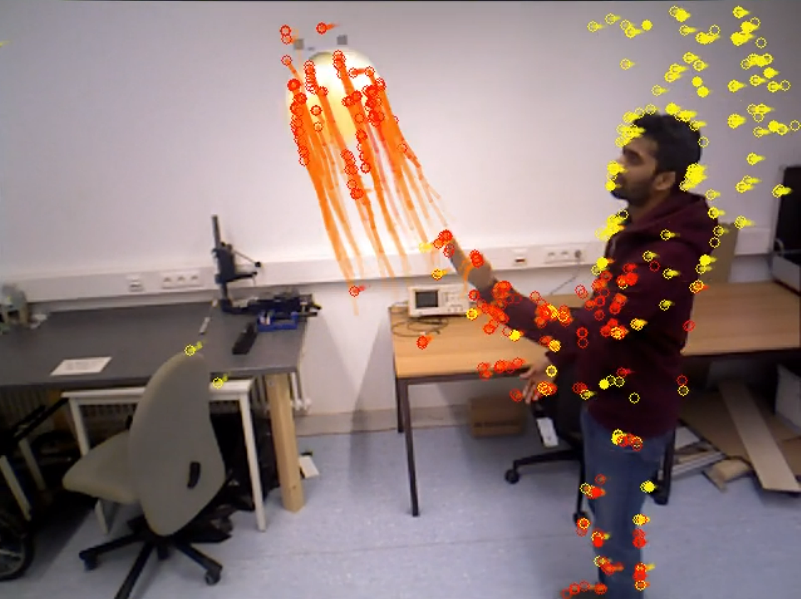}}
\caption{Anchors with high dynamics (in red) and low reliability (in yellow)}
\vspace{-2mm}
\label{fig:leap_point}
\end{figure}

Fig. \ref{fig:leap_point} illustrates the highly dynamic points and points with uncertain dynamic status in the current frame, as identified by the LEAP module. This visualization highlights the ability of LEAP to distinguish between dynamic and uncertain regions, which is crucial for accurate tracking and mapping in dynamic environments.
Unlike other pairwise trackers, LEAP is capable of long-term point tracking across image sequences, demonstrating SOTA performance in dynamic environments. This capability ensures its effectiveness in camera pose estimation and dynamic environment mapping within SLAM systems. Our experimental results further validate this claim.

\subsection{SLAM System}

\subsubsection{Dynamics-aware Tracking} \label{sec:tracking}

In the tracking stage of our SLAM system, we focus on updating the list of tracked points and estimating the camera pose without performing map updates. During initialization, we first use Sobel kernels to obtain the RGB image gradients and apply an average pooling layer to the gradient map to smooth and downsample it. Next, we divide the gradient map into multiple subregions and select the point with the highest image gradient magnitude in each subregion as anchors. The anchors in first frame are selected as tracking anchors.
Then the tracking process begins by leveraging the pre-trained LEAP model to obtain the 2D positions, visibility, dynamic status, and reliability of the tracked points in the current frame. Based on predefined thresholds, we filter out points with low visibility, high dynamic status, low reliability, or insufficient temporal observations, resulting in a stable list of static and trackable points. Specifically, we retain points with visibility \(\mathbf{V}_t \geq \gamma_v\), where \(\gamma_v\) is the visibility threshold, and discard points with dynamic status \(\mathbf{m}_{d,t} \geq \gamma_d\), where \(\gamma_d\) is the dynamic status threshold. Additionally, we keep points with reliability \(\mathbf{\Phi}_t \geq Q(\gamma_u)\), where \(\gamma_u\) is the reliability quantile and \(Q\) is the quantile function, and remove points with fewer than \(\gamma_{track}\) valid observations across the sequence. If the number of tracked points falls below a predefined threshold, we extract new keypoints from \(\mathbf{I}_t\) using the distributed image gradient-based sampling method and add them to the tracking list.

\newcommand{\project}[0]{\pi}
\newcommand{\meanW}[0]{\boldsymbol{\mu}_{W}}
\newcommand{\meanC}[0]{\boldsymbol{\mu}_{C}}
\newcommand{\meanI}[0]{\boldsymbol{\mu}_{I}}
\newcommand{\covW}[0]{\boldsymbol{\Sigma}_{W}}
\newcommand{\covI}[0]{\boldsymbol{\Sigma}_{I}}
\newcommand{\cam}[0]{\boldsymbol{T}}
\newcommand{\camCW}[0]{\cam_{CW}}
\newcommand{\camWC}[0]{\cam_{WC}}
\newcommand{\rotCW}[0]{\boldsymbol{R}_{CW}}
\newcommand{\rotWC}[0]{\boldsymbol{R}_{WC}}
\newcommand{\tauC}[0]{\tau}
\newcommand{\thetaC}[0]{\theta}
\newcommand{\Exp}[0]{\text{Exp}}
\newcommand{\Log}[0]{\text{Log}}
\newcommand{\loss}[0]{\mathcal{L}}
\newcommand{\meddepth}[0]{\tilde{\mathcal{D}}}
\newcommand{\window}[0]{\mathcal{W}}
\newcommand{\kfwindow}[0]{\mathcal{W}_k}
\newcommand{\randwindow}[0]{\mathcal{W}_r}
\newcommand{\pd}[2]{\frac{\partial {#1} }{\partial {#2} }}
\newcommand{\mpd}[2]{\frac{\mathcal{D} {#1}}{\mathcal{D} {#2}}}
\newcommand{\se}[1]{\mathfrak{se}(#1)}
\newcommand{\SE}[1]{\boldsymbol{SE}(#1)}
\newcommand{\identity}[0]{\boldsymbol{I}}
\newcommand{\vectorize}[1]{vec(#1)}
\newcommand{\kron}[0]{\otimes}
\newcommand{\RR}{\mathbb{R}}
\newcommand{\gtimage}[0]{\bar{I}}
\newcommand{\gtdepth}[0]{\bar{D}}
\newcommand{\matJ}[0]{\mathbf{J}}
\newcommand{\matW}[0]{\mathbf{W}}

Then, camera pose estimation is performed on the basis of filtered tracked points, ensuring robustness and accuracy in dynamic environments. We minimize a combination of photometric and geometric residuals, computed exclusively for the tracked points that pass the filtering stage. This approach leverages both color and depth information from the RGB-D camera, while avoiding the influence of unreliable or dynamic points.

The photometric residual measures the difference between the rendered image and the observed image. For the filtered stable and static tracked points, the photometric residual is defined as:
\begin{equation}
    E_{pho} = \left\| I_f(\gaussians, \camCW) - \gtimage_f \right\|_1~,
    \label{eqn:photometric_filtered}
\end{equation}
where \( I_f(\gaussians, \camCW) \) renders the Gaussians \( \gaussians \) from the current camera pose \( \camCW \), and \( \gtimage_f \) is the observed image.

The geometric residual measures the difference between the rendered depth and the observed depth. For the filtered stable and static tracked points, the geometric residual is defined as:
\begin{equation}
    E_{geo} = \left\| D_f(\gaussians, \camCW) - \gtdepth_f \right\|_1~,
    \label{eqn:geometric_residual_filtered}
\end{equation}
where \( D_f(\gaussians, \camCW) \) is the depth rasterization from the current camera pose \( \camCW \), and \( \gtdepth_f \) is the observed depth.

Finally, the camera pose is optimized by minimizing a weighted combination of the photometric and geometric residuals:
\begin{equation}
    E_{total} = \lambda_{pho} E_{pho} + (1 - \lambda_{pho}) E_{geo}~,
    \label{eqn:total_residual_filtered}
\end{equation}
where \( \lambda_{pho} \) is a hyperparameter balancing the contributions of the photometric and geometric residuals. 

By computing residuals exclusively for the filtered tracked points, our camera pose estimation method achieves robustness in dynamic environments while leveraging the full capabilities of the RGB-D camera. This approach ensures accurate and reliable pose estimation, even in the presence of challenging conditions such as occlusions or dynamic objects.

\subsubsection{Keyframing}
\label{sec:keyframing}

Our system adopts a keyframing strategy similar to MonoGS \cite{matsuki2023gaussian}, which involves keyframe selection based on covisibility and relative translation, keyframe management to maintain a small window of non-redundant keyframes, and Gaussian insertion and pruning to refine the scene representation. 
A key distinction in our approach is the use of 4D Gaussians, which incorporate temporal information. When computing covisibility, we set the time parameter to the current frame's timestamp, enabling us to obtain the conditional 3D Gaussians distributed in space at the given time. 

\subsubsection{Dynamics-aware 4DGS Mapping}\label{sec:mapping}
The purpose of mapping is to maintain a coherent spatial-temporal structure and to optimise the newly inserted Gaussians. During mapping, the keyframes in $\kfwindow$ are used to reconstruct currently visible regions. Additionally, two random past keyframes $\randwindow$ are selected per iteration to avoid forgetting the global map.

In our method, we leverage dynamic status obtained from the front-end to distinguish between Gaussians with strong dynamics and those with weak dynamics. For Gaussians with strong dynamics, we apply an isotropic regularisation term that considers both spatial and temporal scaling parameters. For Gaussians with weak dynamics, the isotropic regularisation term only considers spatial scaling parameters. This approach allows us to use fewer Gaussians to represent relatively static parts of the scene while adding more Gaussians in dynamic regions to capture their spatiotemporal features more effectively.
The isotropic regularisation term for Gaussians with strong dynamics is defined as:

\begin{equation}
    E_{iso}^{dynamic} = \sum_{i =1}^{|\gaussians^{dynamic}|} \left\| \vec{S_i^{4D}} - \tilde{\vec{S_i^{4D}}} \cdot \mathbf{1} \right\|_1
\end{equation}
where \(\vec{S}_i^{4D}\) represents the spatiotemporal scaling parameters of a 4D gaussian, and \(\tilde{\vec{S}_i}^{4D}\) is the corresponding mean.
For Gaussians with weak dynamics, the isotropic regularisation term is defined as:
\begin{equation}
    E_{iso}^{static} = \sum_{i =1}^{|\gaussians^{static}|} \left\| \vec{S}_i^{3D} - \tilde{\vec{S}_i}^{3D} \cdot \mathbf{1} \right\|_1
\end{equation}
where only the spatial scaling parameters \(\vec{S}_i^{3D}\) are considered.

\renewcommand{\kfwindow}{\mathcal{K}} 
\renewcommand{\randwindow}{\mathcal{R}} 
Let the union of the keyframes in the current window and the randomly selected ones be \(\window = \kfwindow \cup \randwindow\). For mapping, we solve the following optimisation problem:
\begin{equation}
\small
    \min_{\substack{\camCW^k \in \SE{3}, \gaussians, \\ \forall k \in \window}}
    \sum_{\forall k \in \window} E^{k}_{total} + \lambda_{iso}^{dynamic} E_{iso}^{dynamic} + \lambda_{iso}^{static} E_{iso}^{static}
    ~.\label{eqn:mapping}
\end{equation}

\begin{figure}[!htbp]
\centerline{\includegraphics[width=\columnwidth]{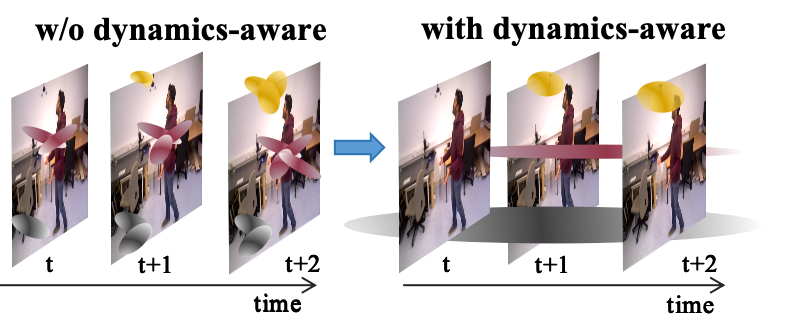}}
\vspace{-2mm}
\caption{Comparison of Gaussian distributions with and without dynamics-aware InfoModule}
\label{fig:dynamicmapping}
\end{figure}
Figure 5 compares Gaussian Distributions with and without the dynamics-aware InfoModule. If we apply the same 4D isotropic regularisation to all Gaussians, multiple 4D Gaussians may exist even in the static parts of the map. However, when we use dynamics-aware isotropic regularisation, the distinction between dynamic and static regions results in a more reasonable distribution of Gaussians along the time axis. In the static parts, the Gaussians have a wide temporal width, while in the dynamic parts, the temporal distribution of Gaussians fits the occurrence of objects well.

\begin{table}[htbp]
    \caption{Results of metric ATE RMSE on several dynamic scene sequences in \textit{BONN} dataset. ``$*$" denotes the version reproduced by NICE-SLAM. ``-" denotes the tracking failures. The metric unit is [cm].}
    \centering
    \resizebox{\columnwidth}{!}{%
    \begin{tabular}{lcccccc}
    \toprule
    Method & \texttt{ball} & \texttt{ball2} & \texttt{ps\_tk} & \texttt{ps\_tk2} & \texttt{ball\_tk} & Avg. \\ 
    \midrule
    ORB-SLAM3~\cite{orbslam3} & 5.8 & 17.7 & 70.7 & 77.9 & 3.1 & 29.8 \\
    ReFusion~\cite{palazzolo2019arxiv} & 17.5 & 25.4 & 28.9 & 46.3 & 30.2 & 27.7 \\
    DROID-VO~\cite{teed2021droid} & 5.4 & 4.6 & 21.4 & 46.0 & 8.9 & 15.4 \\
    \addlinespace 
    \cdashline{1-7} 
    \addlinespace 
    iMAP*\cite{sucar2021imap} & 14.9 & 67.0 & 28.3 & 52.8 & 24.8 & 36.1 \\
    NICE-SLAM\cite{nice_slam} & - & 66.8 & 54.9 & 45.3 & 21.2 & 44.1 \\
    Vox-Fusion\cite{yang2022vox} & 65.7 & 82.1 & 128.6 & 162.2 & 43.9 & 88.4 \\
    Co-SLAM\cite{Wang_2023_CVPR} & 28.8 & 20.6 & 61.0 & 59.1 & 38.3 & 46.3 \\
    ESLAM\cite{Johari_2023_CVPR} & 22.6 & 36.2 & 48.0 & 51.4 & 12.4 & 31.4 \\
    Rodyn-SLAM\cite{jiang2024rodyn} & 7.9 & 11.5 & 14.5 & 13.8 & 13.3 & 12.3 \\
    SplaTAM\cite{keetha2023splatam} & 35.5 & 36.1 & 149.7 & 91.2 & 12.5 & 57.4 \\
    GS-SLAM\cite{yan2023gs} & 37.5 & 26.8 & 46.8 & 50.4 & 31.9 & 33.1 \\
    DG-SLAM \cite{xu2024dgslam} & 3.7 & 4.1 & 4.5 & 6.9 & 10.0 & 5.5 \\ 
    \addlinespace 
    \cdashline{1-7} 
    \addlinespace 
    \textbf{Ours} & \textbf{3.6} & \textbf{3.9} & \textbf{4.5} & \textbf{5.2} & \textbf{8.5} & \textbf{5.1} \\ 
    \bottomrule
    \end{tabular}
    }
    \label{table:bonn-tracking}
\end{table}
\begin{table*}[h]
    \vspace{1.2mm}
    \caption{Camera Tracking and Mapping quality on several dynamic sequences in the \textit{TartanAir-Shibuya} dataset.}
    \vspace{-4mm}
    \begin{center}
    \resizebox{1.0\linewidth}{!}{
    \begin{tabular}{l|cccc|cccc|cccc|cccc|cccc}
    \toprule
     &  \multicolumn{4}{c|}{\texttt{sh01}} & \multicolumn{4}{c|}{\texttt{rc03}} & \multicolumn{4}{c|}{\texttt{rc06}} & \multicolumn{4}{c|}{\texttt{rc07}}  & \multicolumn{4}{c}{\texttt{Avg.}} \\
    
     & ATE↓ & PSNR↑ & SSIM↑ & LPIPS↓ & ATE↓ & PSNR↑ & SSIM↑ & LPIPS↓ & ATE↓ & PSNR↑ & SSIM↑ & LPIPS↓ & ATE↓ & PSNR↑ & SSIM↑ & LPIPS↓ & ATE↓ & PSNR↑ & SSIM↑ & LPIPS↓\\
    \midrule
    SplaTAM \cite{keetha2023splatam} & 64.0 & 12.02 & 0.310 & 0.457 & 52.6 & 11.11 & 0.234 & 0.621 & 82.8 & 13.33 & 0.513 & 0.358 & 93.4 & 13.25 & 0.477  &  0.509 & 73.2 & 12.43 & 0.384 & 0.486 \\
    MonoGS \cite{matsuki2023gaussian}  & 53.0 & 13.91 & 0.303 & 0.576 & 58.8 & 13.21 & 0.241 & 0.638 & 82.8 & 16.63 & 0.516 & 0.404 & 85.6 & 14.74 & 0.489  &  0.534  & 70.1 & 14.62 & 0.387 & 0.538 \\
    \textbf{Ours} &\textbf{3.2} & \textbf{23.48} & \textbf{0.673} & \textbf{0.268} & \textbf{4.1}& \textbf{24.38} & \textbf{0.686} & \textbf{0.267} & \textbf{2.1}& \textbf{21.39} & \textbf{0.793} & \textbf{0.243} & \textbf{5.1} & \textbf{21.73} & \textbf{0.669} & \textbf{0.387} & \textbf{3.6} & \textbf{22.75} & \textbf{0.705} & \textbf{0.291} \\
    \bottomrule 
    \end{tabular}
    }
    \label{table:TartanAir-Shibuya}
    \end{center}
\end{table*}
\setlength{\tabcolsep}{3.5pt}
\begin{table*}[h]
    \vspace{-2mm}
    \caption{Map quality on several dynamic sequences in the \textit{BONN} dataset.}
    \vspace{-4mm}
    \vspace{-2mm}
    \begin{center}
    \footnotesize
    \resizebox{1.0\linewidth}{!}{
    \begin{tabular}{l|ccc|ccc|ccc|ccc|ccc|ccc}
    \toprule
     &  \multicolumn{3}{c|}{\texttt{ball}} & \multicolumn{3}{c|}{\texttt{ball2}} & \multicolumn{3}{c|}{\texttt{ps\_tk}} & \multicolumn{3}{c|}{\texttt{ps\_tk2}}  & \multicolumn{3}{c}{\texttt{ball\_tk}} & \multicolumn{3}{c}{\texttt{Avg.}} \\
    
     & PSNR↑ & SSIM↑ & LPIPS↓  & PSNR↑ & SSIM↑ & LPIPS↓ & PSNR↑ & SSIM↑ & LPIPS↓ & PSNR↑ & SSIM↑ & LPIPS↓ & PSNR↑ & SSIM↑ & LPIPS↓ & PSNR↑ & SSIM↑ & LPIPS↓\\
    \midrule
    SplaTAM \cite{keetha2023splatam} & 17.59 & 0.766 & 0.244 & 16.81 & 0.650 & 0.332 & 18.90 & 0.655 & 0.270 & 17.25 & 0.721 & 0.263 & 15.55 & 0.633 & 0.413 & 17.22 & 0.685 & 0.304 \\
    MonoGS \cite{matsuki2023gaussian}  & 17.72 & 0.712 & 0.478 & 19.44 & 0.747 & 0.367 & 18.8 & 0.736 & 0.399 & 20.01 & 0.755  & 0.375  & 18.89  & 0.623  & 0.272 & 18.97 & 0.715 & 0.378 \\
    \textbf{Ours} & \textbf{27.89} & \textbf{0.857} & \textbf{0.236} & \textbf{29.65} & \textbf{0.839} & \textbf{0.272} & \textbf{27.66} & \textbf{0.832} & \textbf{0.265} & \textbf{31.18} & \textbf{0.876} & \textbf{0.259} & \textbf{27.19} & \textbf{0.865} & \textbf{0.264} & \textbf{28.71} & \textbf{0.854} & \textbf{0.259} \\
    \bottomrule 
    \end{tabular}
    }
    \label{table:rendering}
    \vspace{-2mm}
    \end{center}
\end{table*}
\begin{figure*}[!h]
\centerline{\includegraphics[width=2\columnwidth]{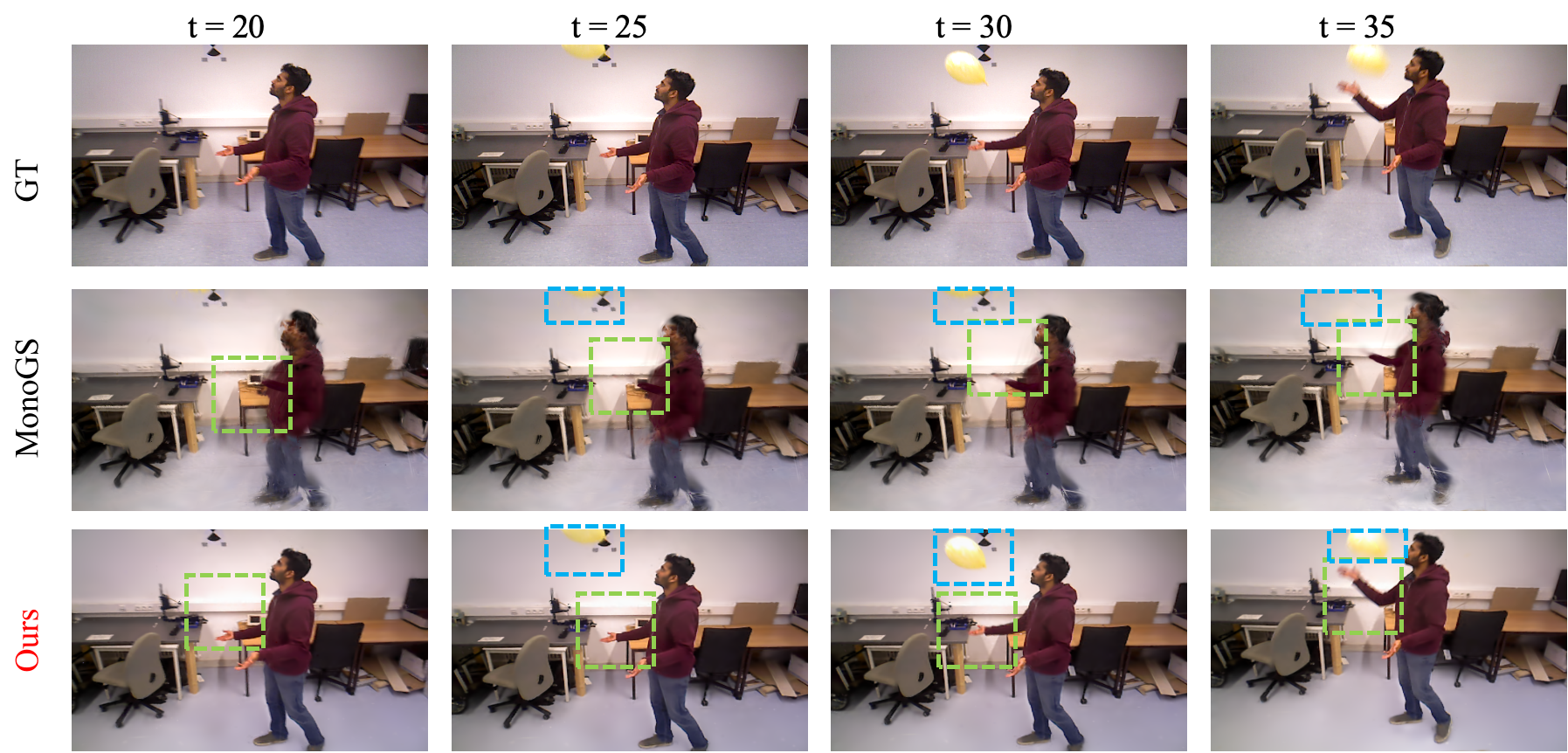}}
\vspace{-2mm}
\caption{Visual comparison of the rendering image on the  \textit{BONN} datasets.}
\label{fig:bonn_render}
\end{figure*}
\begin{figure*}[!h]
\centerline{\includegraphics[width=2\columnwidth]{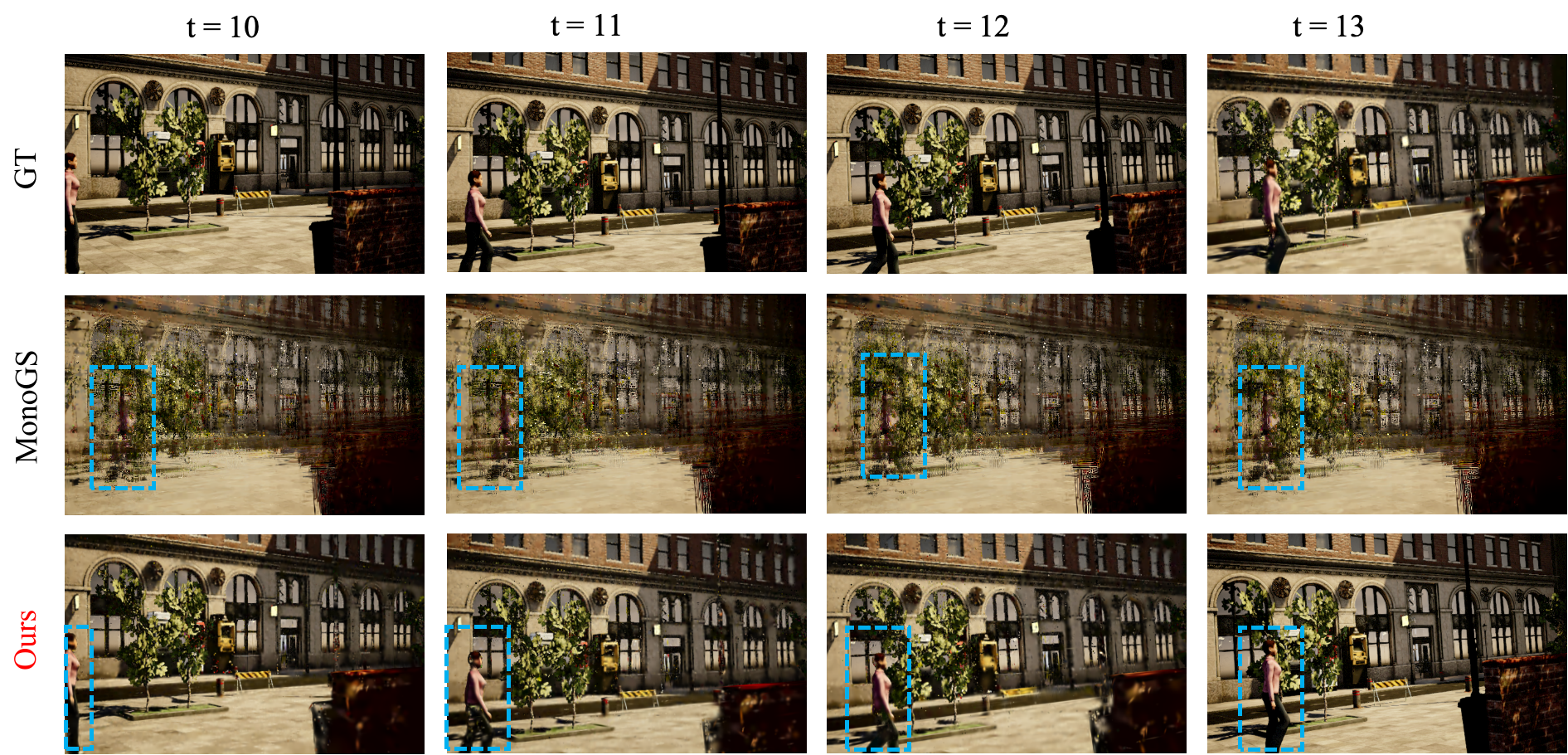}}
\vspace{-2mm}
\caption{Visual comparison of the rendering image on the  \textit{TartanAir-Shibuya} datasets.}
\vspace{-2mm}
\label{fig:shibuya_render}
\end{figure*}

\section{EXPERIMENT}
\subsection{Experimental Setup.}
To evaluate the tracking and mapping performance of our method, we selected the challenging real-world dynamic SLAM datasets BONN datasets \cite{palazzolo2019arxiv} and the more complex TartanAir-Shibuya dataset \cite{qiu2022airdos}. We adopt widely used metrics for camera tracking accuracy and map quality as in \cite{Sandström2023ICCV}. For camera tracking, we report the Root Mean Square Error (RMSE) of the Absolute Trajectory Error (ATE) in centimeters. For map quality, we report photometric rendering metrics, including Peak Signal-to-Noise Ratio (PSNR), Structural Similarity Index (SSIM), and Learned Perceptual Image Patch Similarity (LPIPS).

We run our SLAM on a laptop with an Intel Core i7-13700HX CPU and a single NVIDIA GeForce RTX 4080 GPU. The track filtering parameters are set at $\gamma_v=0.9, \gamma_{d}=0.9, \gamma_u = 0.8, \gamma_{track}=3$. Experimental results show that our algorithm can run at 1.5 FPS on the BONN dataset and 0.9 FPS on the TartanAir-Shibuya dataset, achieving SOTA performance in both tracking and mapping. 
Our primary comparisons are conducted against 3DGS-based methods like DG-SLAM \cite{xu2024dgslam}, MonoGS \cite{matsuki2023gaussian}, GS-SLAM\cite{yan2023gs}, SplaTAM\cite{keetha2023splatam}; additionally, we include comparisons with methods such as  Rodyn-SLAM\cite{jiang2024rodyn}, ESLAM\cite{Johari_2023_CVPR}, Co-SLAM\cite{Wang_2023_CVPR}, Vox-Fusion\cite{yang2022vox}, NICE-SLAM\cite{nice_slam}, iMAP*\cite{sucar2021imap}, ORB-SLAM3~\cite{orbslam3}, ReFusion~\cite{palazzolo2019arxiv} and DROID-VO\cite{teed2021droid}.
It is worth noting that DG-SLAM is a recently introduced method. Since its code has not been open-sourced, the experimental data used in this work are derived from its original paper \cite{xu2024dgslam}.
More implementation details and evaluations will be published at \href{https://github.com/zhicongsun/D4DGS-SLAM}{this link}\footnote{https://github.com/zhicongsun/D4DGS-SLAM}. The specific results are as follows:

\subsection{Evaluation of Camera Tracking Performance}
We evaluate our method on both controlled laboratory scenes (BONN dataset) and complex outdoor dynamic environments (TartanAir-Shibuya). As shown in Table~\ref{table:bonn-tracking}, our approach achieves superior camera pose accuracy compared to state-of-the-art SLAM systems across various dynamic scenarios. Notably, we reduce ATE RMSE by 7.3\% compared to the closest competitor DG-SLAM~\cite{xu2024dgslam} (5.1 cm vs. 5.5 cm), demonstrating the effectiveness of our spatio-temporal Gaussian representation in handling persistent motions. Traditional geometric methods like ORB-SLAM3~\cite{orbslam3} exhibit significant failures in high-dynamic sequences (\textit{e.g.}, 77.9 cm ATE RMSE in \texttt{ps\_tk2}), while neural implicit approaches (NICE-SLAM, Vox-Fusion) suffer from catastrophic tracking failures due to their static scene assumptions.

The advantages of our method become more pronounced in large-scale dynamic environments, as evidenced by the TartanAir-Shibuya results in Table~\ref{table:TartanAir-Shibuya}. Our system achieves a 95.1\% reduction in average ATE compared to SplaTAM~\cite{keetha2023splatam} (3.6 cm vs. 73.2 cm) while simultaneously improving rendering quality metrics, 83.1\% higher PSNR (22.75 vs. 12.43) and 45.9\% better SSIM (0.705 vs. 0.384). This dual improvement stems from our method explicitly models object motions rather than treating them as outliers. Particularly in \texttt{rc06} with rapid camera movements, our method maintains sub-centimeter tracking accuracy (2.1 cm ATE) where baseline Gaussian SLAM systems fail catastrophically (larger than 80 cm ATE RMSE).

These results validate our key design choices. Our approach demonstrates particular strength in scenes with both static and moving objects, where traditional dynamic SLAM methods struggle to balance reconstruction quality with tracking robustness.

\subsection{Evaluation of Mapping Performance}
In terms of mapping performance comparison, we primarily compare with the baseline methods MonoGS~\cite{matsuki2023gaussian} and SplaTAM~\cite{keetha2023splatam}, which are representative 3DGS-based SLAM. The reason for this is that existing NeRF-based SLAM methods and 3DGS-based methods for dynamic scenes do not have the capability to reconstruct a real map; they choose to filter out dynamic objects. Additionally, the latest DG-SLAM~\cite{yan2023gs} code has not been open-sourced yet, so we are temporarily unable to test its mapping performance. However, since DG-SLAM~\cite{yan2023gs}, like MonoGS~\cite{matsuki2023gaussian} and SplaTAM~\cite{keetha2023splatam}, essentially uses 3DGS for mapping, comparing our proposed dynamic map representation method with these static 3DGS methods can still reveal the advantages and disadvantages of our approach.

As shown in Table \ref{table:rendering}, our method achieves 28.71 PSNR, 0.854 SSIM, and 0.259 LPIPS on average, outperforming SplaTAM and MonoGS by significant margins (e.g. +10.5 PSNR and -0.15 LPIPS). The superiority is consistent across all sequences involving moving objects like ball, demonstrating 4DGS’s ability to model spatio-temporal variations without ghosting artifacts.  
Similar trends hold for the TartanAir-Shibuya dataset (Table 2): our approach attains 22.75 PSNR and 0.705 SSIM, surpassing baselines by +8 PSNR and +0.3 SSIM. Notably, the LPIPS score (0.291) is 45\% lower than SplaTAM, indicating superior perceptual consistency.

We visualize rendered scenes from our method and 3DGS-based MonoGS on BONN and TartanAir-Shibuya, respectively. 
As shown in Fig. \ref{fig:bonn_render}, our method effectively reconstructs the moving hand (highlighted by the green box) and the balloon (highlighted by the yellow box). In contrast, MonoGS exhibits delayed reconstruction of the moving hand and fails to reconstruct the newly appeared balloon.
As shown in the Fig. \ref{fig:shibuya_render}, MonoGS exhibit severe ghosting artifacts, such as semi-transparent duplicates of moving objects, which significantly degrade the quality of the reconstructed scene. In contrast, our method effectively captures dynamic features (e.g., the trajectory of a walking person) while preserving static structures with high fidelity.
Furthermore, our results demonstrate smooth transitions between frames (left-to-right in the figures), ensuring temporal consistency in the reconstructed scene. On the other hand, MonoGS suffers from flickering artifacts, which are primarily caused by misestimated Gaussian positions due to dynamic occlusions. This highlights the robustness of our approach in handling dynamic environments compared to traditional 3DGS-based techniques.

\subsection{Ablation study}
The ablation study on the ball scene in the \textit{BONN} dataset demonstrates the effectiveness of our proposed method. Without dynamics-aware InfoModule, the system exhibits significantly higher ATE (27.9 cm) and lower PSNR (20.23), SSIM (0.790), and LPIPS (0.371), highlighting the critical role of dynamics-aware InfoModule in improving tracking accuracy and reconstruction quality. Without 4DGS, the ATE improves to 7.2 cm, but PSNR (18.23), SSIM (0.645), and LPIPS (0.327) remain suboptimal, indicating 4DGS's importance for high-fidelity reconstruction. Our full method achieves the best performance, with an ATE of 3.6 cm, PSNR of 27.89, SSIM of 0.857, and LPIPS of 0.236, confirming the synergistic benefits of integrating LEAP and 4DGS for dynamic scene understanding.
\begin{table}[!h]
    \footnotesize 
    \vspace{-2mm}
    \caption{Ablation study on the ball scene in the \textit{BONN} dataset. D-aware means the dynamics-aware InfoModule}
    \centering
    \vspace{-2mm}
      \begin{tabular}{p{1.6cm}p{1.6cm}p{1cm}p{1cm}p{1cm}}
    \toprule
     & ATE RMSE↓ & PSNR↑ & SSIM↑ & LPIPS↓ \\ 
    \midrule
    w/o D-aware & 27.9 & 20.23 & 0.790 & 0.371 \\
    w/o 4DGS & 7.2 & 18.23 & 0.645 & 0.327\\
    \textbf{Ours} & \textbf{3.6} & \textbf{27.89}  & \textbf{0.857} & \textbf{0.236} \\ 
    \bottomrule
    \end{tabular}
    \label{table:ablation}
\end{table}

\section{CONCLUSIONS}
In this paper, we present D4DGS-SLAM, the first SLAM method based on dynamic map representation and dynamics-aware module. Experimental results on dynamic datasets demonstrate that our method effectively captures the dynamic characteristics of the scene, outperforming state-of-the-art algorithms in both camera tracking accuracy and mapping quality.
In addition to introducing 4DGS into map representation for the first time, we also provide a framework that integrates dynamic information into 4D Gaussian optimization. This will allow our algorithm to improve as dynamic point tracking methods advance. 
Nonetheless, this work serves as a preliminary investigation, with many aspects warranting further study and the potential to reveal additional challenges. For example, tracking performance may degrade under heavy occlusion by dynamic objects, due to inherent difficulties in distinguishing between dynamic and static scene components. Moreover, the current reliance on a large number of Gaussians to represent scenes restricts the applicability of our algorithm on resource-constrained platforms, such as small robots or drones. Future work will aim to explore these and other limitations in greater depth, and to develop effective solutions accordingly.





\bibliographystyle{IEEEtranS}
\bibliography{D4DGS-SLAM/main}

\end{document}